\documentclass[11pt]{article}

\usepackage[a4paper,margin=1in]{geometry}
\usepackage{graphicx}
\usepackage{amsmath}
\usepackage{amssymb}
\usepackage{booktabs}
\usepackage{multirow}
\usepackage{xcolor}
\usepackage{microtype}
\usepackage{authblk}
\usepackage[hidelinks]{hyperref}

\renewcommand{\Affilfont}{\small}
\makeatletter
\renewcommand{\AB@affilsepx}{\protect\qquad\protect\Affilfont}
\makeatother

\begin{document}

\title{CoMVS-GS: Collaborative Multi-View Stereo and 3D Gaussian Splatting for Surface Reconstruction}
\author[1]{Shihan Chen}
\author[1]{Junjing Zhang}
\author[1]{Qingsong Yan}
\author[1]{Haibing Liu}
\author[2]{Haofan Ren}
\author[1]{Fei Deng}
\affil[1]{Wuhan University}
\affil[2]{Hangzhou Dianzi University}
\date{}

\maketitle

\begin{abstract}
3D Gaussian Splatting enables efficient novel view synthesis, but accurate mesh reconstruction remains difficult in weakly observed and occluded regions, where Gaussian primitives may grow into unstable or geometrically inconsistent structures. We propose \textbf{CoMVS-GS}, a general surface reconstruction framework that combines Multi-View Stereo with Gaussian splatting. CoMVS-GS initializes Gaussian primitives from dense multi-view stereo points with pre-flattened scales and normal-aligned orientations, providing stronger geometric priors than sparse structure-from-motion initialization and reducing ambiguity during early optimization. It further introduces PatchMatch-3DGS Mutual Supervision, where Gaussian-rendered depths and normals initialize PatchMatch refinement, and refined PatchMatch depths supervise Gaussian optimization to improve weakly constrained geometry. For surface extraction, CoMVS-GS replaces truncated signed distance field voxel fusion with a Delaunay graph-cut meshing pipeline, reducing sensitivity to voxel resolution while preserving visibility-consistent surface evidence. Experiments on DTU, GauU-Scene V2, and MatrixCity show that CoMVS-GS remains competitive on object-level reconstruction and improves geometric accuracy and mesh compactness in outdoor scenes while maintaining high rendering quality.
\end{abstract}

\noindent\textbf{Keywords:} 3D Gaussian Splatting; Surface Reconstruction; Multi-View Stereo; PatchMatch; Delaunay Graph-Cut Meshing.

\section{Introduction}

\begin{figure}[t]
\centering
\includegraphics[width=\textwidth]{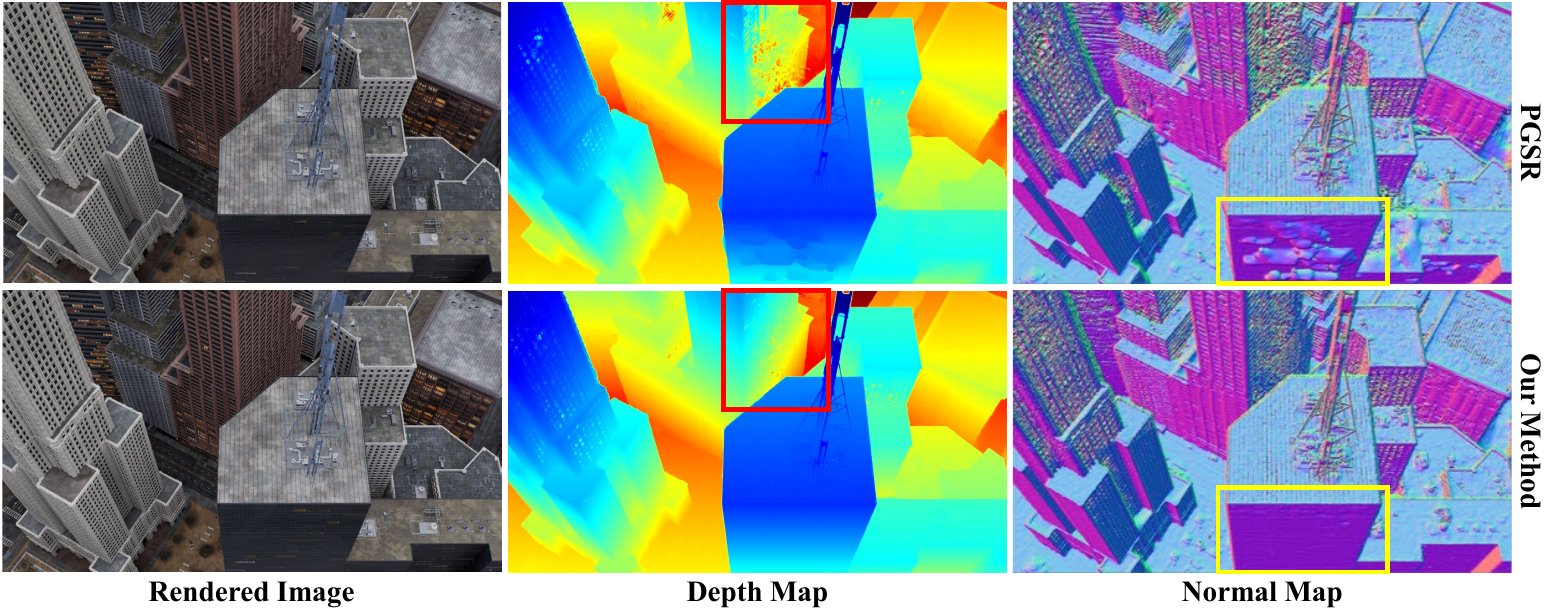}
\caption{Qualitative comparison between PGSR and CoMVS-GS in rendered images, depth maps, and normal maps. CoMVS-GS produces flatter geometry in weakly constrained regions while preserving fine scene details.}
\label{fig:Figure_1}
\end{figure}

Reconstructing accurate 3D surfaces from multi-view images is a fundamental problem in computer vision. Classical Structure from Motion (SfM) and Multi-View Stereo (MVS) pipelines~\cite{schonberger2016sfm,schonberger2016pixelwise} recover geometry by matching image evidence across views, and can produce metrically reliable point clouds and meshes when sufficient texture and overlap are available. In parallel, neural rendering methods have shifted the field toward optimizing scene representations through differentiable rendering. Among them, 3D Gaussian Splatting (3DGS)~\cite{kerbl20233d} has become particularly attractive because it represents scenes with explicit Gaussian primitives and achieves fast training and real-time rendering. However, the original 3DGS formulation is optimized mainly for photometric view synthesis rather than surface reconstruction. Its primitives may therefore reproduce training views well while forming noisy, floating, or geometrically inconsistent structures in 3D.

Recent Gaussian-based surface reconstruction methods attempt to narrow this gap by making Gaussian primitives more surface-aware. SuGaR~\cite{guedon2023sugar}, 2DGS~\cite{huang20242dgs}, Gaussian Surfels~\cite{dai2024gaussian}, GOF~\cite{yu2024gof}, Trim3DGS~\cite{fan2024trim3dgs}, and RaDe-GS~\cite{zhang2026rade} introduce different geometric priors or depth rasterization strategies for extracting surfaces from Gaussian representations. PGSR~\cite{chen2024pgsr} further improves surface quality by flattening Gaussians into planar primitives and imposing multi-view photometric and reprojection constraints. These advances show that 3DGS can be adapted from a rendering representation into a geometry reconstruction framework. Nevertheless, robust mesh reconstruction remains difficult in weakly observed, heavily occluded, and geometrically under-constrained regions, especially in outdoor scenes with complex visibility, as illustrated in Fig.~\ref{fig:Figure_1}.

Most 3DGS-based methods initialize Gaussians from sparse SfM point clouds~\cite{kerbl20233d,guedon2023sugar,huang20242dgs,dai2024gaussian,chen2024pgsr}. While sufficient for radiance-field optimization, these sparse points leave large spatial voids in regions with few reliable feature tracks, forcing geometry and surface orientation to be recovered mainly through later optimization. Classical MVS can instead generate dense point clouds with surface normal estimates before Gaussian optimization begins~\cite{schonberger2016pixelwise,openmvs2020}, providing a stronger geometric prior for sparsely initialized and weakly observed regions.

Geometry supervision during optimization is also often weak or noisy. PGSR-style multi-view regularization relies on photometric and reprojection consistency between a reference view and selected source views~\cite{chen2024pgsr}, but distance- or direction-based neighbors may have limited actual overlap and inconsistent visibility near occlusion boundaries under irregular trajectories. Gaussian optimization therefore benefits most directly from explicit MVS depth refinement, with overlap-aware source-view filtering used as an auxiliary strategy to make multi-view regularization more reliable.

Mesh extraction introduces another source of instability. Many Gaussian reconstruction pipelines render depth maps from optimized Gaussians, fuse them into a TSDF volume~\cite{curless1996tsdf}, and extract a surface with Marching Cubes~\cite{lorensen1987marching,huang20242dgs,chen2024pgsr,yu2024gof}. This strategy is sensitive to voxel resolution: coarse grids over-smooth fine structures, whereas fine grids increase memory consumption and can expose holes or fragmented surfaces when rendered depths are inconsistent. The issue is especially prominent in outdoor reconstruction, although the underlying discretization sensitivity is not limited to large-scale scenes.

To address these challenges, we propose \textbf{CoMVS-GS}, a general MVS-enhanced Gaussian surface reconstruction framework for both object-level and outdoor scenes. Our key idea is to use MVS not as a separate final reconstruction pipeline, but as a source of dense initialization, depth-level supervision, and visibility-consistent meshing. Specifically, MVS dense points initialize Gaussian primitives; PatchMatch-3DGS Mutual Supervision uses rendered depths and normals to initialize PatchMatch and refined PatchMatch depths to supervise 3DGS; and the final surface is recovered with depth fusion, Delaunay graph-cut reconstruction, and mesh refinement rather than TSDF voxel fusion~\cite{jancosek2011multi,vu2012high}. Inspired by common MVS practice, we also use SfM co-visibility with camera baseline and viewing direction to filter PGSR-style source views.

Our main contributions are summarized as follows:
\begin{itemize}
    \item We introduce a pre-flattened and normal-aligned Gaussian initialization strategy using MVS dense point clouds. This approach explicitly embeds geometric priors into the primitives, reducing ambiguous normal recovery in standard isotropic initialization.
    \item We develop PatchMatch-3DGS Mutual Supervision, coupling PatchMatch depth refinement with Gaussian optimization to provide explicit depth-level supervision.
    \item We employ a Delaunay graph-cut meshing pipeline for TSDF-free surface extraction, reducing sensitivity to voxel resolution and improving scalability in outdoor scenes.
\end{itemize}

\section{Related Work}

\subsection{Neural and Gaussian-Based Surface Reconstruction}
Neural radiance fields optimize continuous scene representations from posed images. NeRF~\cite{mildenhall2020nerf} introduced differentiable volume rendering, while VolSDF~\cite{yariv2021volsdf}, NeuS~\cite{wang2021neus}, and Neuralangelo~\cite{li2023neuralangelo} improve geometry with signed distance fields or multi-resolution encodings. These implicit methods provide strong modeling capacity, but usually require dense ray sampling and lengthy optimization.

3D Gaussian Splatting (3DGS)~\cite{kerbl20233d} replaces ray marching with efficient Gaussian rasterization. Since standard 3DGS lacks an explicit surface, later methods impose geometric priors for mesh extraction. SuGaR~\cite{guedon2023sugar} aligns Gaussians with local surfaces, while 2DGS~\cite{huang20242dgs} and Gaussian Surfels~\cite{dai2024gaussian} use flattened primitives. GOF~\cite{yu2024gof} extracts surfaces from opacity fields, 3DGSR~\cite{lyu20243dgsr} adds an implicit surface branch, and RaDe-GS~\cite{zhang2026rade} improves depth rasterization. Their geometry still depends heavily on initialization, regularization, and meshing. PGSR~\cite{chen2024pgsr} further flattens Gaussians and adds multi-view photometric and geometric consistency. Despite this progress, Gaussian reconstruction still faces a tension between appearance fitting and reliable geometry in weakly observed regions.

\subsection{Multi-View Stereo and PatchMatch}
Classical photogrammetric pipelines estimate camera poses with SfM and recover dense geometry with MVS. COLMAP~\cite{schonberger2016sfm,schonberger2016pixelwise} combines robust SfM with pixelwise view selection for unstructured MVS. Semi-Global Matching~\cite{hirschmuller2008stereo} and PatchMatch stereo~\cite{barnes2009patchmatch,bleyer2011patchmatch} are influential dense matching paradigms; the latter estimates per-view depth and normal hypotheses from multi-view matching.

PatchMatch-based MVS improves efficiency through plane propagation, view selection, and consistency filtering. Gipuma~\cite{galliani2015massively} accelerates normal diffusion on GPUs, ACMM~\cite{xu2019acmm} strengthens multi-scale consistency, and TAPA-MVS~\cite{romanoni2019tapa} targets textureless regions. OpenMVS~\cite{openmvs2020} provides a practical pipeline with depth estimation, fusion, Delaunay graph-cut meshing, and mesh refinement. However, MVS remains sensitive to repetitive, or weakly textured surfaces. We use PatchMatch as an explicit geometric supervisor for Gaussian optimization, rather than treating MVS only as separate preprocessing or final reconstruction.

\subsection{Mesh Extraction from Depth Maps}

Volumetric fusion integrates depth observations into a Truncated Signed Distance Function (TSDF)~\cite{curless1996tsdf} and extracts an iso-surface with Marching Cubes~\cite{lorensen1987marching}; it is widely used in libraries such as Open3D~\cite{zhou2018open3d}. Many 3DGS methods also adopt TSDF fusion over rendered depths~\cite{huang20242dgs,chen2024pgsr,yu2024gof}. However, TSDF fusion is sensitive to voxel resolution: coarse grids suppress details, whereas fine grids increase memory use and may expose holes.

Point-based and graph-based alternatives operate directly on samples and visibility. Poisson reconstruction~\cite{kazhdan2006poisson} estimates an implicit indicator from oriented points. Delaunay graph-cut reconstruction builds a tetrahedral complex from samples and camera rays, then infers inside/outside labels through visibility-aware optimization~\cite{labatut2009robust,jancosek2011multi}. This strategy has been integrated into high-resolution MVS pipelines~\cite{vu2012high}. We use a Delaunay graph-cut meshing pipeline~\cite{jancosek2011multi,vu2012high} to convert depth maps refined by PatchMatch-3DGS Mutual Supervision into a mesh, reducing the sensitivity to TSDF voxel resolution.

\section{Method}

\begin{figure}[t]
\centering
\includegraphics[width=\textwidth]{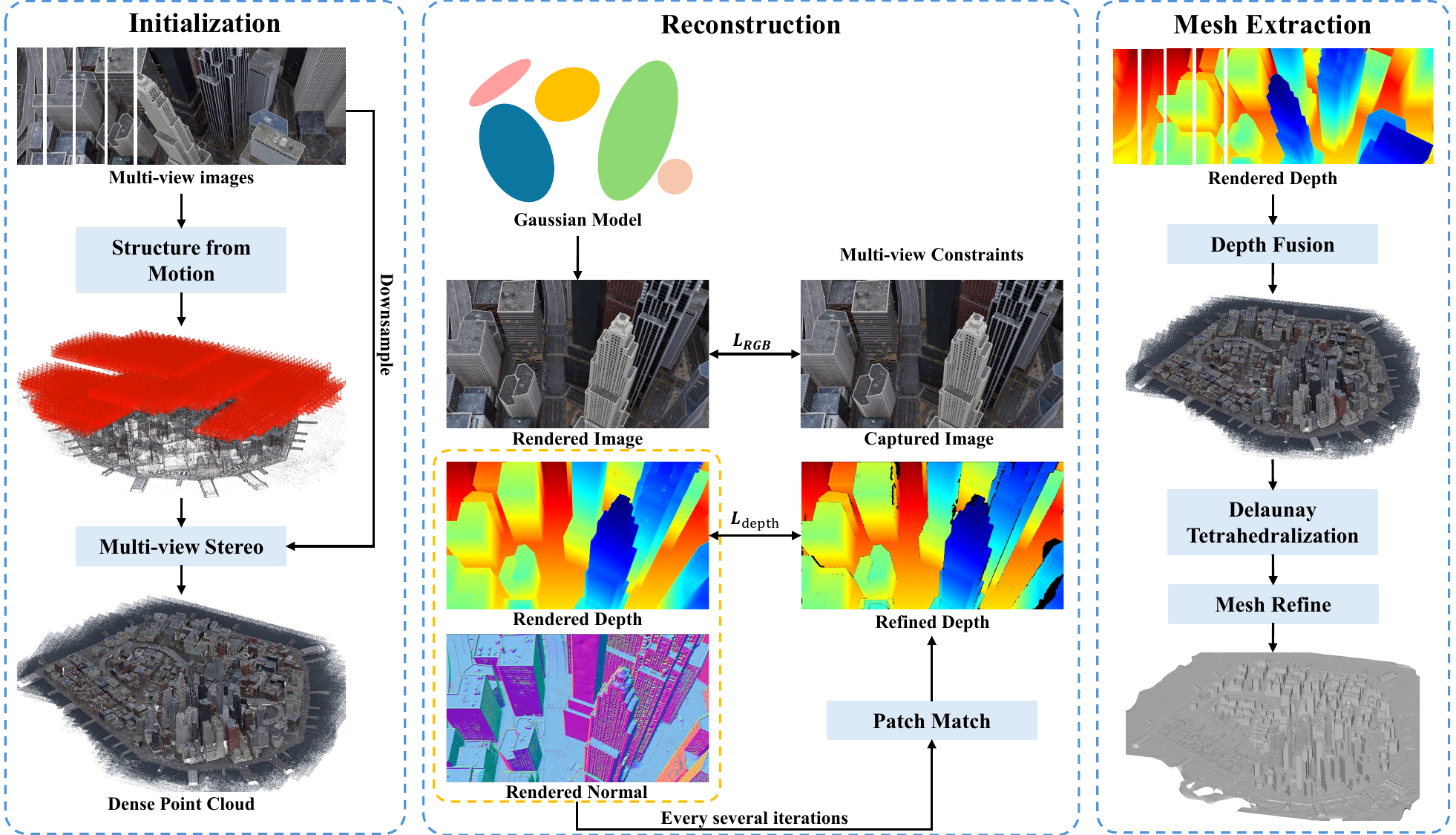}
\caption{Overview of CoMVS-GS. (a) MVS dense points provide pre-flattened and normal-aligned Gaussian initialization without requiring a precomputed mesh. (b) PatchMatch-3DGS Mutual Supervision refines Gaussian geometry, with MVS-style filtering used for PGSR-style source views. (c) TSDF-free surface extraction through a Delaunay graph-cut meshing pipeline.}
\label{fig:pipeline}
\end{figure}

Given posed images, CoMVS-GS integrates PGSR-style surface-aware optimization with explicit MVS geometry and visibility-consistent meshing, as illustrated in Fig.~\ref{fig:pipeline}. It first initializes Gaussians from MVS dense points with pre-flattened scales and MVS-derived normal alignment, providing dense and surface-aware geometry. PatchMatch-3DGS Mutual Supervision then couples Gaussian optimization with MVS depth refinement. For the PGSR-style photometric and reprojection losses, we use an MVS-style source-view filter based on sparse-point co-visibility, camera baseline, and viewing direction. Finally, the optimized depth maps are fused into surface samples and meshed with Delaunay graph cuts, avoiding the voxel-resolution trade-off of TSDF fusion.

\subsection{Preliminaries: Planar Gaussian Splatting}
\label{sec:prelim}

We build on the planar Gaussian representation of PGSR~\cite{chen2024pgsr}. A scene is represented by Gaussian primitives $\mathcal{G}=\{G_i\}_{i=1}^{N}$, each with a center $\boldsymbol{\mu}_i$, opacity $o_i$, anisotropic covariance $\boldsymbol{\Sigma}_i$, and spherical-harmonic color coefficients. For pixel $p$, visible Gaussians are depth-sorted and alpha-blended:
\begin{equation}
    \hat{\mathbf{C}}(p)=\sum_{i\in\mathcal{R}(p)} T_i(p)\alpha_i(p)\mathbf{c}_i,
    \quad
    T_i(p)=\prod_{j<i}\bigl(1-\alpha_j(p)\bigr),
    \label{eq:alpha_blend}
\end{equation}
where $\alpha_i(p)$ is the projected opacity and $T_i(p)$ is the accumulated transmittance.

PGSR flattens each Gaussian along its shortest scale axis, making it a local planar surface element. With camera-space plane normal $\mathbf{n}_i^c$ and signed distance $d_i^c$, the normal map and camera-to-plane distance map are rendered with the same blending weights:
\begin{equation}
    \hat{\mathbf{N}}(p)=\sum_{i\in\mathcal{R}(p)} T_i(p)\alpha_i(p)\mathbf{n}_i^c,
    \quad
    \hat{D}(p)=\sum_{i\in\mathcal{R}(p)} T_i(p)\alpha_i(p)d_i^c.
    \label{eq:pgsr_normal_distance}
\end{equation}
Given ray direction $\mathbf{r}(p)=\mathbf{K}^{-1}\tilde{p}$, the rendered depth is obtained by intersecting the ray with the blended plane:
\begin{equation}
    \hat{z}(p)=\frac{\hat{D}(p)}{\hat{\mathbf{N}}(p)^{\top}\mathbf{r}(p)}.
    \label{eq:pgsr_depth}
\end{equation}

\subsection{Dense Point Cloud Initialization}
\label{sec:init}

While conventional planar Gaussian methods initialize primitives as isotropic spheres and rely on unguided optimization to recover surface normals, we leverage the dense MVS point cloud $\mathcal{P}_{mvs} = \{(\mathbf{x}_i, \mathbf{c}_i, \mathbf{n}_i)\}_{i=1}^{M}$ to introduce a pre-flattened and normal-aligned initialization. This strategy bridges spatial gaps left by sparse SfM points and embeds geometric priors into the primitives from the first iteration.

Specifically, each Gaussian is initialized as a flattened planar primitive aligned with the local MVS surface estimate:
\begin{equation}
    \boldsymbol{\mu}_i = \mathbf{x}_i,\quad
    \mathbf{c}_i^{SH}=\operatorname{RGB2SH}(\mathbf{c}_i),\quad
    \mathbf{R}_i=\operatorname{AlignZ}(\mathbf{n}_i),\quad
    \mathbf{S}_i=\operatorname{diag}(s_{xy}, s_{xy}, \epsilon).
    \label{eq:mvs_init}
\end{equation}

This geometric initialization is implemented through two steps: \textbf{Normal-Aligned Rotation} $\mathbf{R}_i$: We establish a local orthogonal basis using the MVS-derived normal $\mathbf{n}_i$. The rotation matrix $\mathbf{R}_i$ aligns the local $Z$-axis, i.e., the shortest axis of the planar Gaussian, with $\mathbf{n}_i$. \textbf{Pre-flattened Scale} $\mathbf{S}_i$: Instead of standard isotropic scaling, we flatten the initial Gaussian by setting the scale along the normal direction to a small thickness $\epsilon$. The planar extent $s_{xy}$ is adaptively computed from the local neighborhood $\mathcal{N}_i$.

By initializing Gaussians with flattened shapes and MVS-derived orientations, we reduce the reliance on ambiguous normal recovery during early optimization. This provides a better-conditioned starting point and suppresses floating artifacts in weakly observed areas.

\subsection{PatchMatch-3DGS Mutual Supervision}
\label{sec:mutual}

\subsubsection{Mutual Depth Refinement.}
PatchMatch and 3DGS provide complementary geometric signals. PatchMatch uses multi-view matching to produce explicit depth hypotheses, but it can be sensitive to poor initialization and local matching ambiguities. Gaussian optimization provides dense rendered depth and normal maps, yet these maps may drift in weakly constrained regions when only photometric and PGSR-style regularization are used. We therefore couple the two sources of geometry in a closed loop.

During training, we periodically render the current Gaussian depth $\hat{z}$ and normal $\hat{\mathbf{N}}$ using Eqs.~\eqref{eq:pgsr_normal_distance}--\eqref{eq:pgsr_depth}. These maps initialize the plane hypotheses in PatchMatch, giving the MVS a geometry-aware starting point. PatchMatch then refines the depth maps by multi-view propagation, matching, and geometric consistency filtering, producing refined depths $z_{pm}$ and a valid-pixel mask $\mathcal{M}_{pm}$. The refined depth is used as pseudo ground truth for Gaussian optimization:
\begin{equation}
    \mathcal{L}_{pm}=
    \frac{1}{|\mathcal{M}_{pm}|}
    \sum_{p\in\mathcal{M}_{pm}}
    \rho\bigl(\hat{z}(p)-z_{pm}(p)\bigr),
    \label{eq:pm_loss}
\end{equation}
where $\rho(\cdot)$ is a robust $\ell_1$ penalty. This mutual supervision makes the Gaussian depth more metrically grounded, while the rendered Gaussian geometry improves the stability of subsequent PatchMatch refinement.

\begin{figure}[t]
\centering
\includegraphics[width=\textwidth]{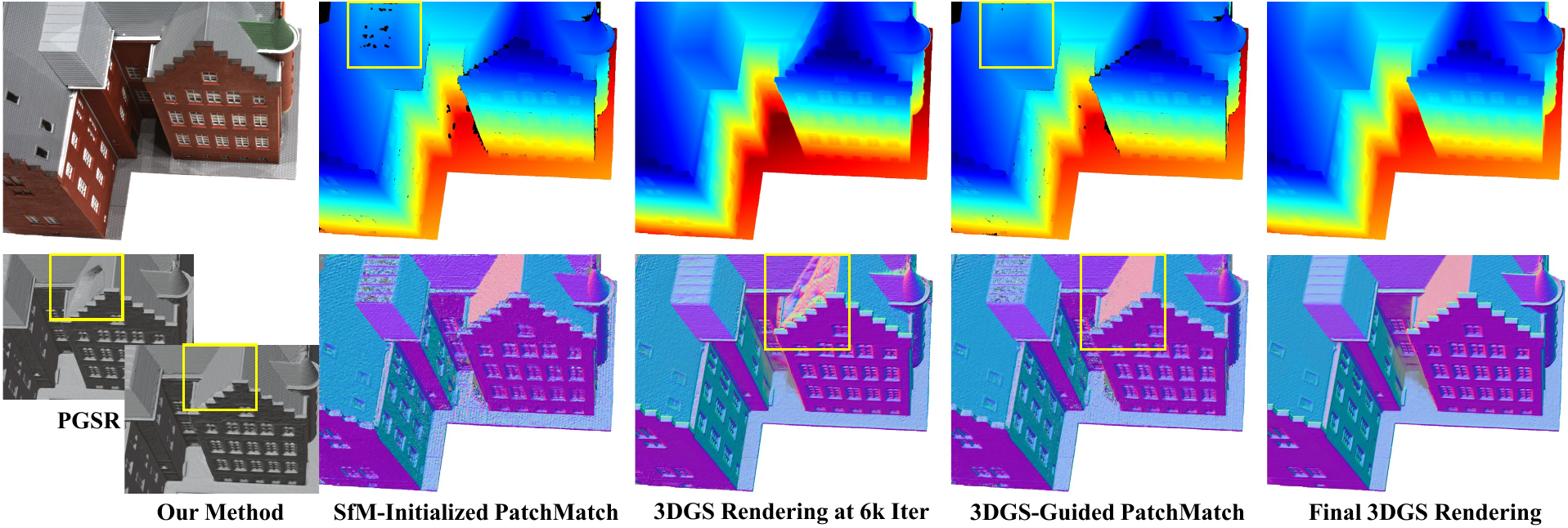}
\caption{Illustration of PatchMatch-3DGS Mutual Supervision.}
\label{fig:Illustration_of_Patch_Match}
\end{figure}

Fig.~\ref{fig:Illustration_of_Patch_Match} illustrates this interaction. PatchMatch initialized only from the SfM result produces incomplete depth maps with visible holes. After 6k iterations, the Gaussian model already renders a more coherent depth and normal estimate, which provides a stronger initialization for PatchMatch. With this guidance, PatchMatch fills missing regions and corrects noisy normals on weakly observed geometry; the refined depth is then used to supervise subsequent Gaussian optimization. The final model consequently renders complete and accurate depth and normal maps.

\subsubsection{Source-View Filtering for Multi-View Regularization.}
Inspired by common MVS practice, we rank source views for the PGSR-style photometric and reprojection losses using sparse-point overlap and camera-pair geometry. For each image $I_i$, let $\mathcal{Q}_i$ be the set of SfM points visible in that image. We first compute the co-visibility ratio between two images as
\begin{equation}
    s_{ij}=\frac{|\mathcal{Q}_i\cap\mathcal{Q}_j|}
    {\min(|\mathcal{Q}_i|,|\mathcal{Q}_j|)}.
    \label{eq:covis_score}
\end{equation}
For a reference image $I_r$, the source-view set is selected as
\begin{equation}
    \mathcal{S}(I_r)=
    \operatorname{TopK}\{I_j\mid
    \Phi(s_{rj}, b_{rj}, \theta_{rj})>\tau_{view}\},
    \label{eq:covis_neighbors}
\end{equation}
where $b_{rj}$ and $\theta_{rj}$ denote the camera-baseline and viewing-direction consistency, and $\Phi(\cdot)$ is a ranking score combining overlap and camera geometry. The photometric constraint $\mathcal{L}_{pc}^{filt}$ and geometric constraints $\mathcal{L}_{gc}^{filt}$ are then computed over $I_s\in\mathcal{S}(I_r)$. Compared with using camera distance or viewing direction alone, adding sparse-point co-visibility better reflects actual image overlap under irregular camera trajectories, so the selected source views provide more useful PGSR-style supervision. We emphasize that this is an auxiliary filter for differentiable Gaussian regularization and does not modify PatchMatch's internal source-view selection.

Our training objective augments the PGSR losses with PatchMatch depth supervision and the filtered multi-view regularization:
\begin{equation}
    \mathcal{L}=
    \mathcal{L}_{rgb}
    +\lambda_s\mathcal{L}_{s}
    +\lambda_n\mathcal{L}_{dn}
    +\lambda_{pm}\mathcal{L}_{pm}
    +\lambda_{pc}\mathcal{L}_{pc}^{filt}
    +\lambda_{gc}\mathcal{L}_{gc}^{filt},
    \label{eq:total_loss}
\end{equation}
where $\mathcal{L}_{s}$ is the Gaussian flattening regularizer inherited from PGSR, $\mathcal{L}_{pm}$ is the PatchMatch depth supervision term, and $\mathcal{L}_{pc}^{filt}$ and $\mathcal{L}_{gc}^{filt}$ denote the PGSR multi-view losses computed with filtered source views.

\subsection{Delaunay Graph-Cut Meshing Pipeline}
\label{sec:mesh}

After optimization, we render depth and normal maps from the final Gaussian model and fuse geometrically consistent depth pixels into surface samples. Following standard MVS filtering~\cite{shen2013accurate}, each reference depth is checked against neighboring views using reprojection, relative-depth, and normal-angle consistency before being unprojected into a fused point cloud $\mathcal{P}_{fused}$.

Instead of accumulating all samples in a TSDF volume, we adopt visibility-consistent Delaunay graph-cut reconstruction~\cite{jancosek2011multi}. A 3D Delaunay tetrahedralization is built from $\mathcal{P}_{fused}$ and camera centers, and camera rays provide free-space and occupied-space evidence for graph-cut labeling. The mesh is extracted from facets that separate tetrahedra with different labels and is then refined with image-based photometric and visibility constraints~\cite{vu2012high}. This sample-and-ray formulation operates on depth-derived samples and camera-ray visibility rather than a dense voxel grid, reducing the memory growth and voxel-size sensitivity of high-resolution TSDF fusion.

\section{Experiments}

\subsection{Experimental Setup}

\textbf{Datasets and Implementation Details.} We evaluate CoMVS-GS on both object-level and outdoor reconstruction. For object-level geometry, we use the standard DTU benchmark~\cite{jensen2014dtu}, which provides calibrated images and laser-scanned reference surfaces. For outdoor scenes, we construct three representative subsets: \textit{Sub-Village} with 301 images from the \textit{Village} scene and \textit{Sub-Technology College} with 600 images from \textbf{GauU-Scene V2}~\cite{xiong2024gauuscenev2}, and \textit{Sub-Small City} with 986 images from \textbf{MatrixCity}~\cite{li2023matrixcity}. These subsets avoid the excessive cost of full city-scale processing while retaining diverse land-cover, building layouts, thin structures, weakly textured surfaces, repeated facades, and occlusions. For all reconstruction experiments, input images are resized so that the longer side is 1600 pixels. All experiments are conducted on a GPU server equipped with an NVIDIA GeForce RTX 4090D.

\textbf{Optimization Details.} We optimize for 30,000 iterations with Adam. The appearance term is $\mathcal{L}_{rgb}=0.8\mathcal{L}_{1}+0.2(1-\operatorname{SSIM})$, and the loss weights in Eq.~\eqref{eq:total_loss} are $\lambda_s=100$, $\lambda_n=0.015$, $\lambda_{pm}=0.3$, $\lambda_{pc}=0.15$, and $\lambda_{gc}=0.03$. The appearance and flattening terms are active throughout training; $\mathcal{L}_{pm}$ starts at iteration 3,000, and $\mathcal{L}_{dn}$, $\mathcal{L}_{pc}^{filt}$, and $\mathcal{L}_{gc}^{filt}$ start at iteration 7,000. For $\mathcal{L}_{pc}^{filt}$, we use $7\times7$ patches, at most 102,400 samples, and a one-pixel reprojection threshold. The position learning rate decays from $4\times10^{-5}$ to $1.6\times10^{-6}$ and the scale learning rate is 0.001; other optimizer settings follow 3DGS~\cite{kerbl20233d}. Densification and pruning run every 200 iterations from iteration 1,000 to 15,000.

\textbf{PatchMatch Configuration.} We run the CUDA depth-map estimator in OpenMVS 2.3.0~\cite{openmvs2020} at iterations 3,000, 6,000, 9,000, 12,000, and 15,000, initializing each run with the current Gaussian-rendered depth and normal maps. The refined depths supervise the Gaussian model until the next update and remain fixed after the final call. We use resolution level 0 with a maximum image dimension of 1,600 pixels, two sub-resolution levels, up to eight source views, four standard and two geometric-consistency iterations, and a $9\times9$ matching window. The depth, normal, and photometric-cost thresholds are 0.01, $25^{\circ}$, and $1-\mathrm{NCC}=0.9$, respectively; fusion requires three agreeing views, post-processing flag 7 is enabled, and the pseudo-depth mask retains confidence values above 0.7.

\textbf{Baselines \& Metrics.} On DTU, we adopt the standard 15-scan evaluation setting used in Geometry-Grounded Gaussian Splatting~\cite{zhang2026geometrygrounded}. The compared methods include OpenMVS~\cite{openmvs2020} as a conventional photogrammetric baseline, implicit neural reconstruction methods, i.e., NeRF~\cite{mildenhall2020nerf}, VolSDF~\cite{yariv2021volsdf}, NeuS~\cite{wang2021neus}, and Neuralangelo~\cite{li2023neuralangelo}, as well as explicit Gaussian-based methods, including 3DGS~\cite{kerbl20233d}, 2DGS~\cite{huang20242dgs}, GOF~\cite{yu2024gof}, 3DGSR~\cite{lyu20243dgsr}, RaDe-GS~\cite{zhang2026rade}, PGSR~\cite{chen2024pgsr}, and Geometry-Grounded Gaussian Splatting~\cite{zhang2026geometrygrounded}. We report Chamfer Distance (CD) for each scan. No runtime comparison is reported on DTU, since our focus on this benchmark is geometric accuracy.

For \textit{Sub-Village}, \textit{Sub-Technology College}, and \textit{Sub-Small City}, we compare against methods that represent different reconstruction paradigms: Neuralangelo~\cite{li2023neuralangelo} for implicit SDF reconstruction, 2DGS~\cite{huang20242dgs}, RaDe-GS~\cite{zhang2026rade}, and PGSR~\cite{chen2024pgsr} for explicit 3DGS-based surface reconstruction, CityGS-X~\cite{gao2025citygsx} for scalable city-level Gaussian reconstruction, and OpenMVS~\cite{openmvs2020} for a conventional photogrammetric pipeline. Rendering quality is evaluated on held-out images with PSNR, SSIM~\cite{wang2004ssim}, and LPIPS~\cite{zhang2018lpips}. We use the one-sided reference-point-to-mesh distance adopted in the outdoor evaluation protocol~\cite{chen20253d}, and report MAE and RMSE. Unlike symmetric Chamfer Distance on DTU, this metric avoids penalizing reconstructed regions that are visible in the images but fall outside the reference LiDAR or simulated geometry.

\subsection{Reconstruction Comparison}

\begin{table}[t]
\centering
\caption{Quantitative comparison on the DTU dataset. We report Chamfer Distance, with the best and second-best results highlighted in bold and underlined.}
\label{tab:dtu}
\setlength{\tabcolsep}{2pt}
\resizebox{\textwidth}{!}{%
\begin{tabular}{lcccccccccccccccc}
\toprule
Method & 24 & 37 & 40 & 55 & 63 & 65 & 69 & 83 & 97 & 105 & 106 & 110 & 114 & 118 & 122 & Mean \\
\midrule
NeRF~\cite{mildenhall2020nerf} & 1.90 & 1.60 & 1.85 & 0.58 & 2.28 & 1.27 & 1.47 & 1.67 & 2.05 & 1.07 & 0.88 & 2.53 & 1.06 & 1.15 & 0.96 & 1.49 \\
VolSDF~\cite{yariv2021volsdf} & 1.14 & 1.26 & 0.81 & 0.49 & 1.25 & 0.70 & 0.72 & 1.29 & 1.18 & 0.70 & 0.66 & 1.08 & 0.42 & 0.61 & 0.55 & 0.86 \\
NeuS~\cite{wang2021neus} & 1.00 & 1.37 & 0.93 & 0.43 & 1.10 & 0.65 & 0.57 & 1.48 & 1.09 & 0.83 & 0.52 & 1.20 & 0.35 & 0.49 & 0.54 & 0.84 \\
Neuralangelo~\cite{li2023neuralangelo} & \underline{0.37} & 0.72 & 0.35 & 0.35 & 0.87 & 0.54 & 0.53 & 1.29 & 0.97 & 0.73 & 0.47 & 0.74 & 0.32 & 0.41 & 0.43 & 0.61 \\
\midrule
3DGS~\cite{kerbl20233d} & 2.14 & 1.53 & 2.08 & 1.68 & 3.49 & 2.21 & 1.43 & 2.07 & 2.22 & 1.75 & 1.79 & 2.55 & 1.53 & 1.52 & 1.50 & 1.96 \\
2DGS~\cite{huang20242dgs} & 0.48 & 0.91 & 0.39 & 0.39 & 1.01 & 0.83 & 0.81 & 1.36 & 1.27 & 0.76 & 0.70 & 1.40 & 0.40 & 0.76 & 0.52 & 0.80 \\
GOF~\cite{yu2024gof} & 0.50 & 0.82 & 0.37 & 0.37 & 1.12 & 0.74 & 0.73 & 1.18 & 1.29 & 0.68 & 0.77 & 0.90 & 0.42 & 0.66 & 0.49 & 0.74 \\
3DGSR~\cite{lyu20243dgsr} & 0.44 & 0.96 & 0.40 & 0.36 & 1.02 & 0.80 & 0.64 & 1.20 & 1.08 & 0.97 & 0.54 & 0.72 & 0.37 & 0.52 & 0.42 & 0.70 \\
RaDe-GS~\cite{zhang2026rade} & 0.43 & 0.75 & 0.35 & 0.37 & 0.81 & 0.74 & 0.74 & 1.19 & 1.20 & 0.65 & 0.61 & 0.84 & 0.35 & 0.66 & 0.46 & 0.68 \\
PGSR~\cite{chen2024pgsr} & \underline{0.37} & \underline{0.54} & 0.44 & 0.37 & \underline{0.78} & 0.57 & 0.49 & 1.06 & \textbf{0.63} & \underline{0.59} & 0.47 & \underline{0.50} & 0.30 & 0.37 & \underline{0.34} & \underline{0.52} \\
GGGS~\cite{zhang2026geometrygrounded} & \underline{0.37} & \textbf{0.50} & \textbf{0.27} & \textbf{0.31} & 0.81 & \textbf{0.43} & \textbf{0.42} & 1.05 & \underline{0.64} & \textbf{0.52} & \textbf{0.32} & 0.58 & 0.30 & \textbf{0.31} & \textbf{0.33} & \textbf{0.48} \\
\midrule
OpenMVS~\cite{openmvs2020} & 0.44 & 0.69 & \underline{0.28} & \underline{0.32} & 0.80 & 0.67 & 0.50 & \textbf{0.72} & 0.72 & 0.68 & \underline{0.41} & 0.57 & \underline{0.29} & 0.39 & 0.49 & 0.53 \\
\midrule
CoMVS-GS & \textbf{0.36} & 0.55 & \underline{0.28} & \textbf{0.31} & \textbf{0.49} & \underline{0.51} & \underline{0.43} & \underline{0.93} & 0.87 & 0.78 & 0.44 & \textbf{0.36} & \textbf{0.27} & \underline{0.34} & 0.35 & \textbf{0.48} \\
\bottomrule
\end{tabular}
}
\end{table}

\begin{table}[t]
\centering
\caption{Quantitative comparison on the Sub-Village, Sub-Technology College, and Sub-Small City. We report visual metrics, geometric metrics, and the average number of mesh faces. Dashes indicate unavailable or inapplicable metrics.}
\label{tab:outdoor}
\setlength{\tabcolsep}{5pt}
\resizebox{\textwidth}{!}{%
\begin{tabular}{lcccccc}
\toprule
\multirow{2}{*}{Method} & \multicolumn{3}{c}{Visual Metrics} & \multicolumn{2}{c}{Geometric Metrics} & \multirow{2}{*}{\#Faces ($\times10^6$)} \\
\cmidrule(lr){2-4}\cmidrule(lr){5-6}
 & PSNR $\uparrow$ & SSIM $\uparrow$ & LPIPS $\downarrow$ & MAE (m) $\downarrow$ & RMSE (m) $\downarrow$ & \\
\midrule
Neuralangelo~\cite{li2023neuralangelo} & 24.650 & 0.686 & 0.404 & 0.816 & 1.473 & 33.063 \\
2DGS~\cite{huang20242dgs} & 25.569 & 0.799 & 0.277 & 0.853 & 1.408 & 35.954 \\
RaDe-GS~\cite{zhang2026rade} & 25.481 & 0.845 & 0.198 & 0.680 & 1.330 & 39.065 \\
PGSR~\cite{chen2024pgsr} & \textbf{28.370} & \textbf{0.886} & 0.174 & 0.670 & 1.284 & 50.169 \\
CityGS-X~\cite{gao2025citygsx} & 27.783 & \underline{0.883} & \textbf{0.150} & 0.755 & 1.466 & 146.591 \\
\midrule
OpenMVS~\cite{openmvs2020} & -- & -- & -- & \underline{0.607} & \underline{1.210} & 46.561 \\
\midrule
CoMVS-GS & \underline{27.813} & \underline{0.883} & \underline{0.173} & \textbf{0.603} & \textbf{1.133} & 8.762 \\
\bottomrule
\end{tabular}
}
\end{table}

\begin{figure}[t]
\centering
\includegraphics[width=\textwidth]{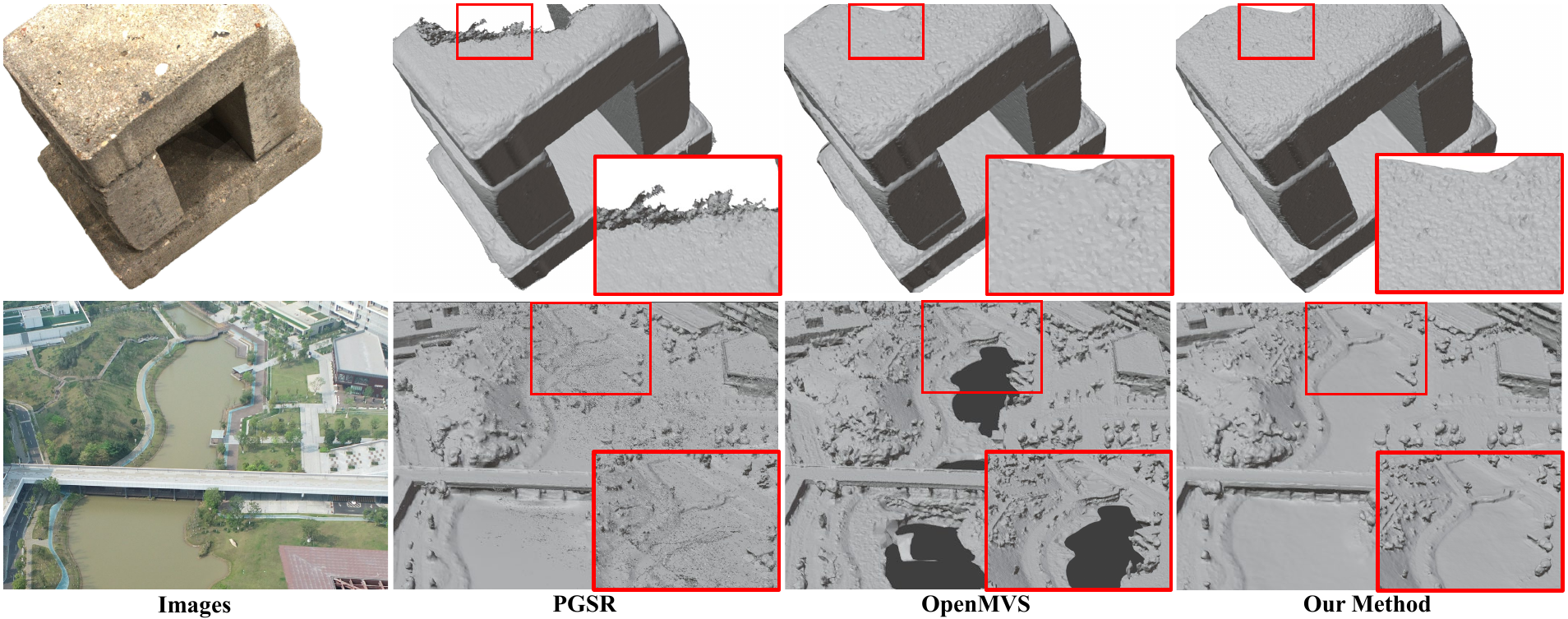}
\caption{Qualitative comparison on DTU scan 40 and Sub-Technology College. From left to right, each row shows the input image, PGSR, OpenMVS, and CoMVS-GS.}
\label{fig:qualitative_comparison}
\end{figure}

Table~\ref{tab:dtu} reports the DTU Chamfer Distance results under the standard 15-scan protocol. CoMVS-GS obtains an average CD of 0.48, which is comparable to the strongest geometry-oriented baselines and clearly improves over PGSR and OpenMVS. The first row of Fig.~\ref{fig:qualitative_comparison} shows the qualitative comparison on DTU scan 40. PGSR fails to fully reconstruct weakly observed boundary regions and produces visible floaters, leading to a smaller surface coverage than OpenMVS and CoMVS-GS. In contrast, CoMVS-GS reconstructs a complete surface comparable to OpenMVS while preserving the advantages of Gaussian optimization. Fig.~\ref{fig:Illustration_of_Patch_Match} further shows that, around the roof region of a DTU house model, CoMVS-GS avoids the local depression observed in PGSR and produces a flatter surface.

Table~\ref{tab:outdoor} further evaluates reconstruction on outdoor scenes. CoMVS-GS remains close to PGSR and CityGS-X in rendering metrics, ranking second in PSNR and SSIM, while achieving the best geometric accuracy with the lowest MAE and RMSE. This contrast suggests that high image-space fidelity does not necessarily translate into the most accurate surface geometry, especially in outdoor regions with occlusion and uneven view coverage. The second row of Fig.~\ref{fig:qualitative_comparison} illustrates this behavior on Sub-Technology College. PGSR produces floating fragmented surfaces above the scene due to insufficient geometric constraints, while OpenMVS leaves large holes in weakly textured water regions. CoMVS-GS combines MVS geometric evidence with Gaussian optimization, yielding a cleaner and more complete reconstruction without obvious floating triangles or water-surface holes. 

The last column of Table~\ref{tab:outdoor} also shows that CoMVS-GS produces substantially more compact meshes, with only 8.762M faces on average. TSDF-based extraction typically requires high-resolution voxels to preserve fine structures, which can produce dense, heavy meshes with many faces, especially in outdoor scenes. In contrast, Delaunay graph-cut meshing operates on depth-derived surface samples and camera-ray visibility, allowing CoMVS-GS to preserve reliable geometric evidence while avoiding excessive face growth from dense voxelization.

\subsection{Ablation Studies}

\begin{figure}[t]
\centering
\includegraphics[width=\textwidth]{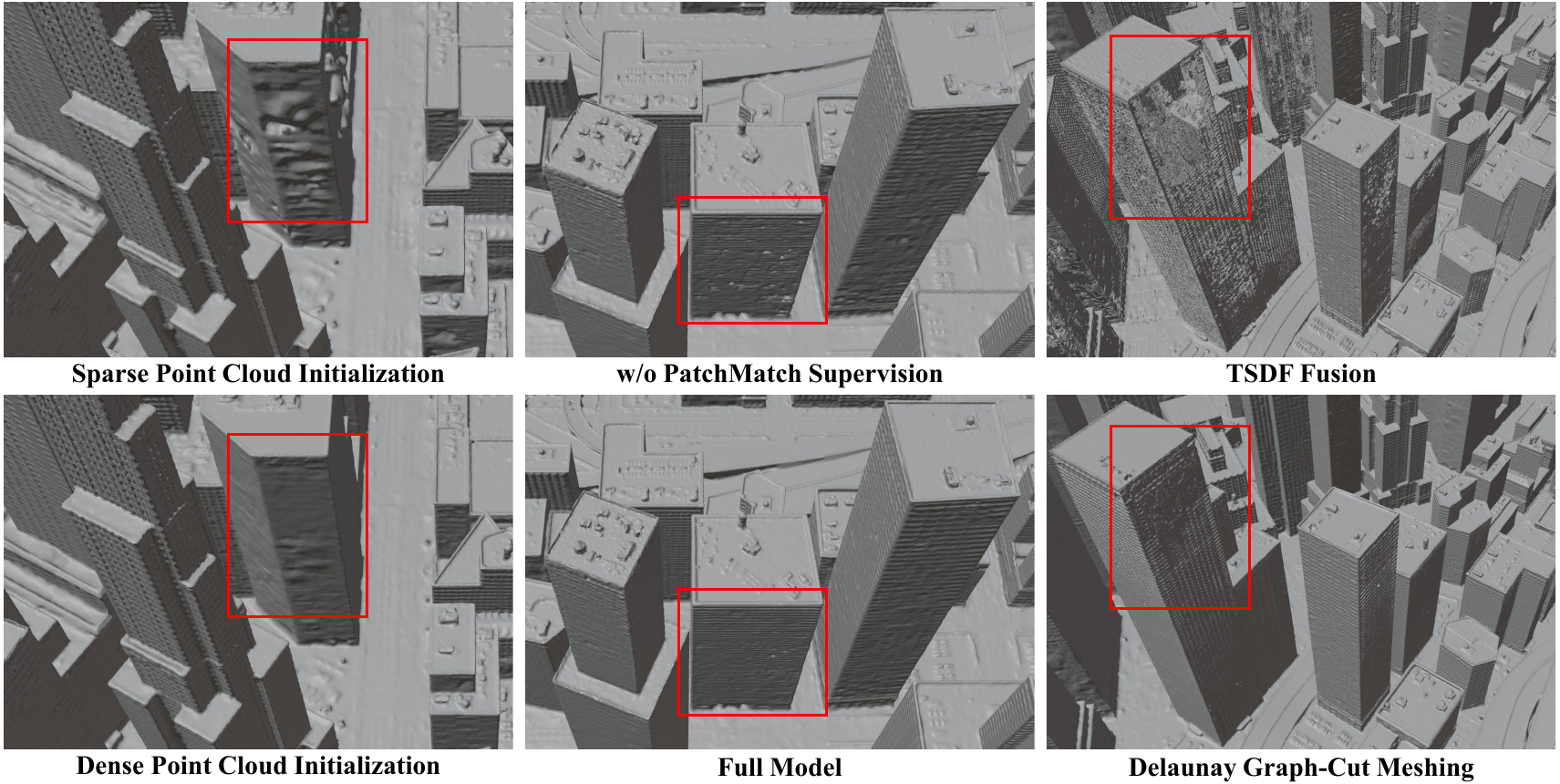}
\caption{Ablation on the Sub-Small City, showing the effects of PatchMatch supervision, Delaunay graph-cut meshing, and dense point cloud initialization.}
\label{fig:Ablation_mesh}
\end{figure}

\textbf{Impact of Dense Point Cloud Initialization.} 
The ablation experiments are conducted on Sub-Small City. Fig.~\ref{fig:Ablation_mesh} shows that initializing 3DGS from dense MVS points with pre-flattened scales and normal-aligned orientations leads to flatter building facades. Sparse SfM initialization provides few reliable anchors in weakly featured or sparsely observed facade regions, making it difficult for Gaussians to converge to the correct surface. Table~\ref{tab:ablation_components} further confirms that dense initialization improves both geometric error metrics.

\textbf{PatchMatch-3DGS Mutual Supervision.} 
PatchMatch-3DGS Mutual Supervision further stabilizes regions with limited observations. Building facades are often viewed at oblique angles and occupy only small image areas, so photometric optimization alone provides insufficient geometric constraints. The proposed mutual supervision supplies persistent depth-level guidance from refined PatchMatch maps, resulting in smoother and more planar building surfaces in Fig.~\ref{fig:Ablation_mesh}. As shown in Table~\ref{tab:ablation_components}, this module improves surface accuracy, although the visual metrics slightly decrease compared with the variant without PatchMatch supervision. This trade-off is consistent with the observation in PGSR~\cite{chen2024pgsr}: stronger geometric constraints restrict the freedom of Gaussians to overfit training views, which may reduce rendering quality but yields more reliable geometry.

\textbf{TSDF vs. Delaunay Graph-Cut Meshing Scalability.} 
We also compare the Delaunay graph-cut meshing pipeline with TSDF fusion. TSDF extraction requires choosing a voxel size that balances detail preservation and completeness. In this experiment, even a relatively fine 0.4m voxel resolution still produces holes and incomplete regions in Fig.~\ref{fig:Ablation_mesh}. Delaunay graph-cut meshing instead reasons over depth-derived surface samples and camera-ray visibility, producing a more complete mesh and achieving the best overall geometric accuracy in Table~\ref{tab:ablation_components}.

\begin{table}[t]
\centering
\caption{Ablation of PatchMatch supervision, Delaunay graph-cut meshing, and dense point cloud initialization. We report visual metrics and geometric metrics.}
\label{tab:ablation_components}
\setlength{\tabcolsep}{5pt}
\resizebox{\textwidth}{!}{%
\begin{tabular}{lccccc}
\toprule
\multirow{2}{*}{Variant} & \multicolumn{3}{c}{Visual Metrics} & \multicolumn{2}{c}{Geometric Metrics} \\
\cmidrule(lr){2-4}\cmidrule(lr){5-6}
 & PSNR $\uparrow$ & SSIM $\uparrow$ & LPIPS $\downarrow$ & MAE (m) $\downarrow$ & RMSE (m) $\downarrow$ \\
\midrule
Base Model & 28.851  & 0.882  & 0.189 & 0.922  & 1.727 \\
TSDF & \underline{28.994} & \underline{0.898} & \textbf{0.157} & \underline{0.543} & 1.161 \\
w/o PatchMatch & \textbf{29.341} & \textbf{0.899} & \underline{0.161} & 0.579 & \underline{1.133} \\
SfM Initialization & 28.860 & 0.882 & 0.189 & 0.599 & 1.165 \\
Full Model & \underline{28.994} & \underline{0.898} & \textbf{0.157} & \textbf{0.512} & \textbf{1.047} \\
\bottomrule
\end{tabular}
}
\end{table}

\section{Conclusion}
We presented CoMVS-GS, an MVS-enhanced 3DGS framework for surface reconstruction across object-level and outdoor scenes. By initializing Gaussians from dense MVS points, coupling PatchMatch depth refinement with Gaussian optimization, and extracting meshes through Delaunay graph cuts, CoMVS-GS improves geometric stability in weakly constrained regions without relying on TSDF voxel fusion. The method uses MVS not as an independent final reconstruction stage, but as dense initialization, iterative depth-level supervision, and visibility-consistent surface evidence. Experiments on DTU, GauU-Scene V2, and MatrixCity show that CoMVS-GS remains competitive on object-level reconstruction and achieves accurate geometry with substantially more compact meshes. The ablation studies further verify that dense initialization, PatchMatch-3DGS Mutual Supervision, and Delaunay graph-cut meshing each contribute to the final reconstruction quality. This highlights the importance of jointly considering initialization, optimization supervision, and mesh extraction in geometry-aware Gaussian reconstruction.

\noindent\textbf{Limitations and Future Work.} CoMVS-GS still inherits PatchMatch's limitations on highly specular or transparent surfaces, where reliable multi-view correspondences are difficult to establish. Periodic PatchMatch refinement also introduces additional computational overhead, especially for high-resolution outdoor scenes. Future work will explore learned matching costs and adaptive refinement schedules to further improve robustness and efficiency.

\bibliographystyle{plain}
\bibliography{references}

\end{document}